\documentclass[10pt,twocolumn,letterpaper]{article}

\usepackage[pagenumbers]{iccv} 
\usepackage{booktabs} 
\usepackage{graphicx}  
\usepackage{caption}   
\usepackage{amsmath}
\usepackage{amssymb}
\usepackage{mathtools}
\usepackage{amsthm}
\usepackage{multirow}

\usepackage{algorithm}
\usepackage{algpseudocode}
\usepackage{xspace}

\newcommand{\modelname}{AnomalyPainter\xspace}
\usepackage{comment}
\usepackage{float}
\usepackage{amssymb}

\definecolor{iccvblue}{rgb}{0.21,0.49,0.74}
\usepackage[pagebackref,breaklinks,colorlinks,allcolors=iccvblue]{hyperref}

\title{\modelname: Vision-Language-Diffusion Synergy for \\ Zero-Shot  Realistic \& Diverse Industrial Anomaly Synthesis}

\author{Zhangyu Lai$^{1}$\footnotemark[2]\quad Yilin Lu$^{1}$\footnotemark[2]\quad Xinyang Li$^{1}$\quad Jianghang Lin$^{1}$\quad Yansong Qu $^{1}$
\\
Liujuan Cao$^{1}$\footnotemark[1]\quad Ming Li$^{2}$\quad Rongrong Ji$^{1}$
\\ \vspace{-0.6em} \\ 
$^1$Key Laboratory of Multimedia Trusted Perception and Efficient Computing,\\
Ministry of Education of China, Xiamen University
\quad $^2$INSPUR DIGI ENT.
\\
}
\begin{document}

\twocolumn[{
\maketitle
\vspace{-3.4em}
\renewcommand\twocolumn[1][]{#1}
\begin{center}
    \centering
    \includegraphics [width=1\textwidth]{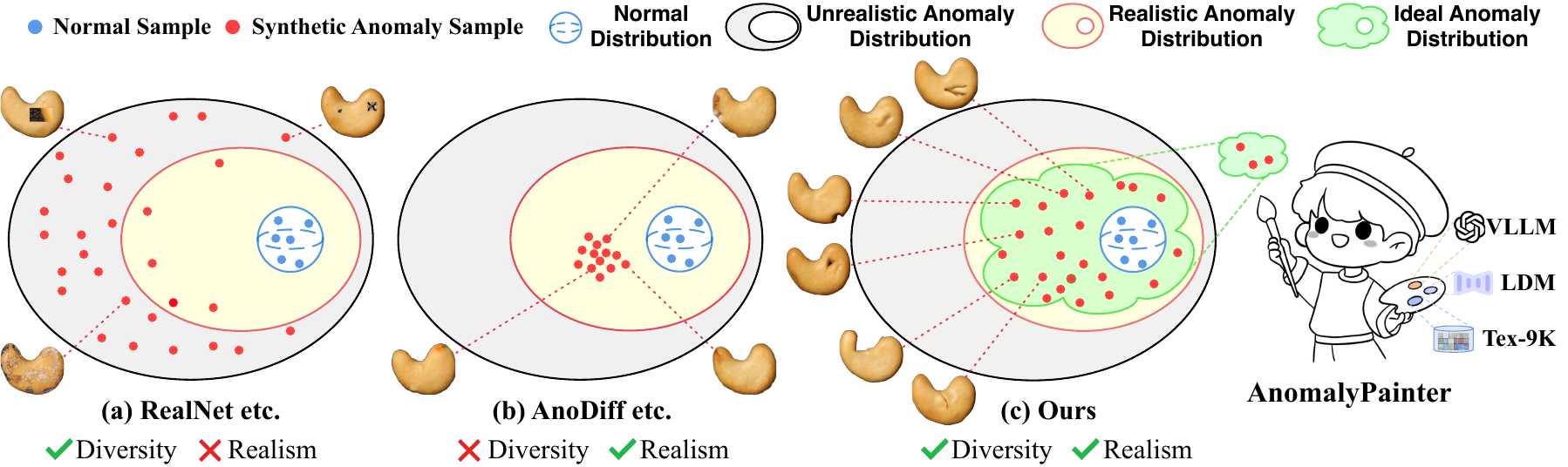}
    \captionof{figure}{
    The blue hypersphere~\cite{liu2024deep} represents the normal sample distribution, realistic anomaly distribution should be close to it, while unrealistic anomaly distribution should be farther.
    Anomaly samples synthesized by different methods exhibit different distributions.
    }
\label{fig:teaser}
\vspace{-2.5pt}
\end{center}
}]
\renewcommand{\thefootnote}{\fnsymbol{footnote}}
\footnotetext[1]{Corresponding Author.}
\footnotetext[2]{Equal Contribution.}

\begin{abstract}
While existing anomaly synthesis methods have made remarkable progress, achieving both realism and diversity in synthesis remains a major obstacle.
To address this, we propose \modelname, a zero-shot framework that breaks the diversity-realism trade-off dilemma through synergizing Vision Language Large Model (VLLM), Latent Diffusion Model (LDM), and our newly introduced texture library Tex-9K. Tex-9K is a professional texture library containing 75 categories and 8,792 texture assets crafted for diverse anomaly synthesis.
Leveraging VLLM’s general knowledge, reasonable anomaly text descriptions are generated for each industrial object and matched with relevant diverse textures from Tex-9K.
These textures then guide the LDM via ControlNet to paint on normal images. 
Furthermore, we introduce Texture-Aware Latent Init to stabilize the natural-image-trained ControlNet for industrial images.
Extensive experiments show that \modelname outperforms existing methods in realism, diversity, and generalization, achieving superior downstream performance. 
\end{abstract}    
\section{Introduction}
\label{sec:intro}
Anomaly detection plays a crucial role in practical applications such as industrial quality control~\cite{liu2024deep} and medical anomaly detection~\cite{huang2024adapting}. 
However, real-world anomaly samples are often exceedingly rare, presenting significant challenges for anomaly detection tasks, including image-level classification and pixel-level segmentation.
Consequently, anomaly synthesis tasks~\cite{hu2024anomalyxfusion,hu2024anomalydiffusion,chen2023easynet,duan2023few,niu2020defect,ojha2021few,zhang2021defect,zhang2024realnet,li2021cutpaste,zavrtanik2021draem} gradually emerge as an advanced technique to expand the dataset of available anomaly samples. 
Depending on whether they employ a limited number of real anomaly samples, existing anomaly synthesis methods can be primarily categorized into zero and few-shot approaches.

Zero-shot methods like CutPaste~\cite{li2021cutpaste}, DRAEM~\cite{zavrtanik2021draem}, and Realnet~\cite{zhang2024realnet}, synthesize diverse anomalies by cropping and pasting patches from existing anomalies or anomaly texture datasets onto normal samples. 
However, this random operation often results in unreasonable anomaly content and abrupt transitions at anomaly boundaries, significantly reducing the realism.
The distribution of zero-shot synthesized anomaly samples, abstractly represented as red scatter points, is similar to that shown in Figure~\ref{fig:teaser} (a). 
These red scatter points tend to fall within the unrealistic anomaly distribution, distant from the feature space of normal samples, which are abstractly represented as blue scatter points.

Few-shot methods like DFMGAN~\cite{duan2023few}, AnoGen~\cite{gui2024anogen}, and AnoDiff~\cite{hu2024anomalydiffusion} use generative models to learn anomaly patterns for generation.
However, the samples synthesized by these methods often overfit the few anomaly samples provided during training, failing to cover the diverse types of anomalies that may appear in real-world objects. 
Similar to Figure~\ref{fig:teaser} (b), the red scatter points representing the few-shot synthesized anomaly samples with similar features, tend to cluster in certain regions in realistic anomaly distribution.

In other words, existing anomaly synthesis methods face the diversity-realism trade-off dilemma. However, we argue that the two are not mutually exclusive.
To achieve both diversity and realism, we propose AnomalyPainter, a zero-shot framework. 
It synergizes the Vision Language Large Model (VLLM) and Latent Diffusion Model (LDM) with our proposed texture library Tex-9K to simulate the formation of real anomalies via texture variations.

Specifically, we construct Tex-9K, a texture library with appropriate texture density, designed for diverse anomaly synthesis. It contains 75 categories and 8,792 professional texture image assets.
Leveraging the extensive general knowledge of VLLM, for each industrial object, it generates reasonable and diverse anomaly text descriptions.
These descriptions retrieve the most relevant textures from Tex-9K, which serve as anomaly pattern conditions to guide the LDM via ControlNet’s~\cite{zhang2023adding} edge-mask control for inpainting normal images.
Since ControlNet is trained on natural images, it sometimes performs unstably on industrial images. 
Texture-Aware Latent Init is then introduced to stabilize ControlNet’s handling, ensuring that normal and texture images blend clearly in the latent space as the initial denoising point, enabling the LDM to achieve precise denoising performance.
In short, our zero-shot AnomalyPainter paints reasonable content (via VLLM) with diverse anomaly patterns (via Tex-9K) on normal images while ensuring soft transitions (via LDM) at anomaly boundaries, ultimately synthesizing diverse and realistic anomaly samples that exhibit an ideal distribution, as shown in Figure~\ref{fig:teaser}(c).

\noindent Our contributions are threefold:
\begin{itemize}
\item We propose \modelname, a zero-shot framework that synergies the general knowledge of VLLM, the high-realism of LDM, and professional assets of Tex-9K, to synthesize realistic and diverse anomaly samples.
\item We propose Tex-9K, a professional texture library containing 75 categories and 8,792 texture assets, designed for broadened texture diversity. Texture-Aware Latent Init is proposed to stabilize ControlNet’s edge-mask control, effectively translating the diverse texture assets provided by Tex-9K into realistic and diverse anomaly samples.
\item  Extensive experiments demonstrate that our synthesized anomaly samples surpass current state-of-the-art (SOTA) synthesis methods and effectively enhance the performance of downstream anomaly detection tasks.
\end{itemize}

\section{Related Work}
\subsection{Anomaly Synthesis} 
Anomaly synthesis has become an essential technique to support anomaly detection, especially when real anomaly samples are scarce. 
Zero-shot methods~\cite{li2021cutpaste,zavrtanik2021draem,chen2023easynet} crop and paste patches from existing anomalies or anomaly texture datasets onto normal samples. 
RealNet~\cite{zhang2024realnet} further employs a generative model to apply noise to normal images and then denoise them to obtain an anomaly dataset, but ultimately pastes part of the anomaly back using a random mask.
While these approaches can produce diverse anomaly samples, the synthesized samples often exhibit significant discrepancies when compared to real-world anomalies.
Few-shot methods~\cite{hu2024anomalydiffusion,hu2024anomalyxfusion,niu2020defect,duan2023few,zhang2021defect,gui2024anogen} use generative models~\cite{abdal2019image2stylegan,rombach2022high} to learn anomaly patterns and generate additional anomaly samples. However, they impose strict constraints on data features, causing the synthesized anomalies to overfit the limited training samples.
\subsection{Vision Language Large Models}
Since the rise of vision-language models (VLMs)~\cite{zhang2024vision}, CLIP~\cite{radford2021learning} has gained significant attention. 
WinCLIP~\cite{jeong2023winclip} first introduces CLIP for assisting anomaly detection, sparking a wave of subsequent advancements~\cite{gu2024anomalygpt,huang2024adapting,cao2024adaclip,zhu2024llms}. However, relatively little exploration has been done in the area of anomaly synthesis.
AnomalyControl~\cite{he2024anomalycontrol} leverages VLM as a cross-modal intermediary, incorporating textual or visual information into the image denoising process to enable controllable anomaly synthesis, similar to our approach. However, its approach requires separate training for each type of control, limiting its generalizability.
More recently, VLMs have been evolving into vision-language large models (VLLMs)~\cite{dong2025internlm,OpenAI2023GPT4V,wang2024qwen2}. Leveraging prior world knowledge, VLLMs can perform complex visual question answering with professional responses.

\begin{figure*}[!t]
  \centering
  \includegraphics[width=1\linewidth]{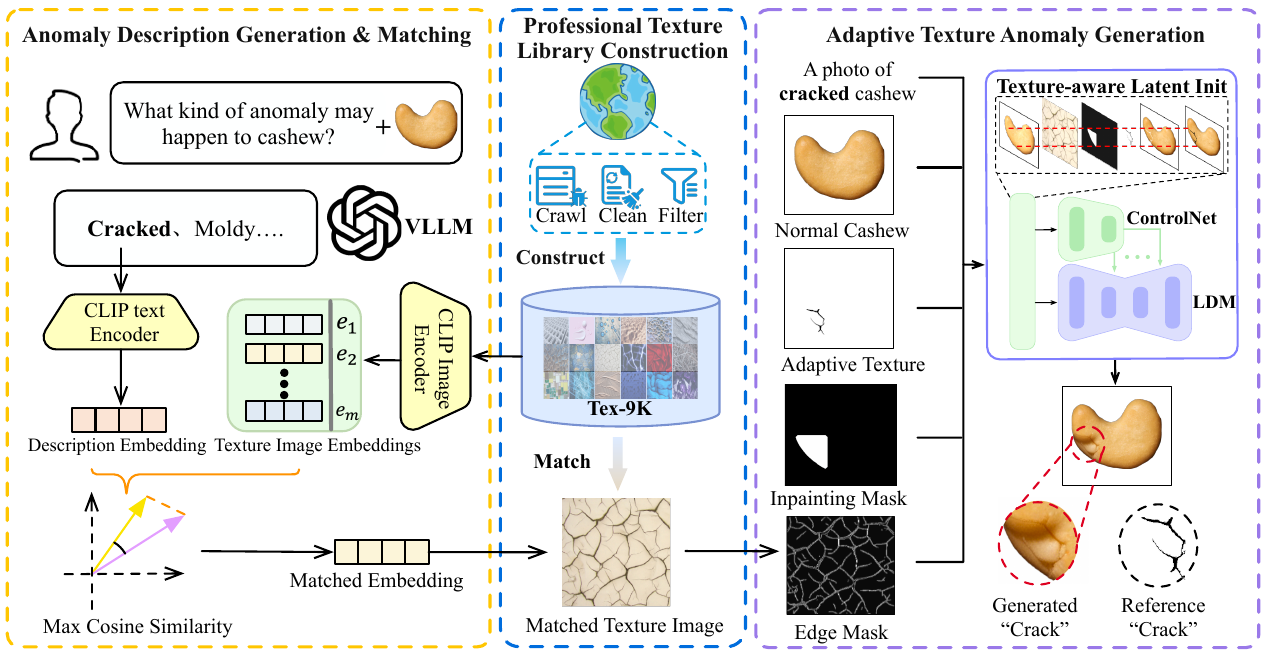}
  \caption{
  \textbf{Overview of \modelname.} Our framework synthesizes diverse and realistic anomaly samples through three main steps: \textbf{Middle:} \textbf{Professional Texture Library Construction} constructs Tex-9K, a texture library with 8,792 texture assets, designed to provide diverse textures crafted for anomaly synthesis. \textbf{Left:} \textbf{Anomaly Description Generation and Matching} utilizes VLLM to generate reasonable anomaly descriptions for each industrial object and matches them with relevant textures from Tex-9K using cosine similarity. \textbf{Right:} \textbf{Adaptive Texture Anomaly Generation} utilizes Texture-Aware Latent Init to stabilize ControlNet’s edge-mask control for LDM's high-realism inpainting, ensuring the seamless integration of relevant textures into normal industrial object images. 
  } 
  \vspace{-10pt}
  \label{fig.2}
\end{figure*}

\subsection{Latent Diffusion Models}
Recent advances in latent diffusion models (LDMs)~\cite{rombach2022high,saharia2022photorealistic,ramesh2022hierarchical,peebles2023scalable,podell2024sdxl}, such as Stable Diffusion~\cite{rombach2022high}, have significantly improved image generation. However, pure text-based control struggles to capture complex scene requirements, leading to the development of various diffusion model plugins~\cite{zhang2023adding,ye2023ip,hu2022lora,zhao2023uni,ruiz2023dreambooth,gal2022image,li2023gligen} for more precise control.
Among these, ControlNet~\cite{zhang2023adding} excels in structural control by integrating additional signals like edge maps, allowing users to specify structures during generation. While many pre-trained ControlNet models are available, they are typically fine-tuned on natural image datasets, leading to instability when applied to industrial objects.
Training-free image composition methods~\cite{lu2023tf,meng2022sdedit,jiang2025pixelman,liu2024training,liu2024training1} improve denoising by guiding attention or blending latents for cross-domain synthesis. However, these methods primarily insert whole objects into images and struggle to blend textures seamlessly with industrial objects.

\section{Method}
Empirically, we consider that the formation of real industrial anomaly samples is usually constrained by the physical properties of objects, which can be understood using the general knowledge of VLLMs.
Under various potential random circumstances, texture variations related to the object may manifest in the image. To effectively simulate this and achieve diverse and realistic anomaly synthesis, we propose \modelname, which is implemented in three key steps: Professional Texture Library Construction, Anomaly Description Generation and Matching, and Adaptive Texture Anomaly Generation. The overview is shown in Figure~\ref{fig.2}.

\subsection{Preliminaries}
\noindent \textbf{Latent Diffusion Models} consist of two key components: an auto-encoder~\cite{kingma2022autoencoding} and a latent denoising network.
The autoencoder establishes a bi-directional mapping from the space of the original data to a low-resolution latent space:
$z = \mathcal{E}(x), x = \mathcal{D}(z),$
where $\mathcal{E}$ and $\mathcal{D}$ are the encoder and decoder respectively. 
The latent denoising network $\epsilon_\theta$ is trained to denoise noisy latent given a specific timestep $t$ and textual prompt embedding $p$.
The diffusion process adopts the standard formulation DDIM~\cite{song2020denoising}, which comprises a forward add-noise diffusion and a backward denoising process. 
During noise addition, the noisy latent representation at a specified timestep $t$ is obtained as: $z_t = \sqrt{\bar{\alpha}_t} \mathcal{E}(x) + \sqrt{1 - \bar{\alpha}_t} \epsilon$, $\bar{\alpha}_t$ is a monotonically decreasing noise schedule and $\epsilon \sim \mathcal{N}(0,1)$ is random noise.
By continuously denoising the random noise $z_T$ with textual prompt embedding $p$ through predicting noise $\epsilon_\theta(z_t,t,p)$, we can derive a fully denoised latent ${z_0}$.
Then, the final clean latent $z_0$ is passed through the latent decoder $\mathcal{D}$ to generate the high-resolution image ${x_0} = \mathcal{D}({z_0})$. 

\subsection{Professional Texture Library Construction} 
\label{sec.4.2}
The motivation for constructing Tex-9K stems from the potential mismatch between our visual intuition for text descriptions and the understanding embedded in VLMs, which are pre-trained on large-scale image-text data from the web. 
For instance, when searching for images of the description ‘cracked’ online, not all results correspond to the clear crack textures required for fine-grained anomaly synthesis. 
Similarly, existing texture libraries, such as DTD~\cite{cimpoi14describing}, often contain overly complex textures that are unsuitable for this purpose.
To smoothly align text descriptions with suitable visual concepts for anomaly synthesis and provide more diverse textures, we expanded existing texture datasets by collecting additional texture images on the Internet and defining 75 commonly used texture categories.

The construction of our texture library can be divided into three parts: 
(a) Crawling: We use a web crawler to collect corresponding images from the Internet legally for each texture category. Data from existing texture datasets are also incorporated into the corresponding categories in our library.
(b) Cleaning: The Canny operator~\cite{canny1986computational} is applied to each image to extract the edge mask, and images with excessively dense or sparse textures edge are automatically discarded.
(c) Filtering: The remaining data is manually screened to retain images with clear textures and remove potentially harmful or inappropriate content.

Ultimately, Tex-9K which retains 8,792 images, serves as the texture library to provide texture assets crafted for anomaly synthesis. See \textit{Appendix.A} for more details. 

\subsection{Anomaly Description Generation \& Matching}
\label{sec.4.3}
Previous zero-shot anomaly synthesis methods randomly paste patches onto normal images, which fail to capture reasonable anomaly types. 
To address this, we pioneer the application of VLLM's general knowledge to enable zero-shot anomaly description simulation, detailed in Alg~\ref{alg1}.

Specifically, let $ O = \{ o_1, o_2, \dots \} $ denote the set of industrial object categories.
We first preprocess the entire Tex-9K by encoding all texture images $\mathcal{X}=\{x_{\text{tex}, 1},x_{tex_2}, \dots, x_{tex_m}\}$ through the CLIP image encoder. 
This generates static Texture Images Embeddings $e_{\mathcal{X}}=\{e_1,e_2, \dots, e_m\}$, which are cached for persistent reuse.
For each possible industrial object  $o_i \in O$, we select a normal image  $x_{N}^{o_i}$  corresponding to the object ${o_i}$, then pose a carefully designed question $Q_{o_i}$ for $o_i$ with the prompt template.
We will detail the template in \textit{Appendix.B}. 
By querying VLLMs with $Q_{o_i}$ and $x_{N}^{o_i}$, we obtain the anomaly description answer $A_{o_i}=\{d_1,d_2, \dots, d_k\}$, where each $d \in A_{o_i}$ is a description.
The CLIP text encoder then encodes $d$ into description embedding $e_{d}$.
After computing cosine similarity between $e_{d}$ and each $e_i \in e_{\mathcal{X}}$, the max one is taken as matched embedding $e_{\text{match}}$, $e_{\text{match}} \in e_\mathcal{X}$. 
The corresponding $x_{\text{match}} \in \mathcal{X}$ is then used as the matched texture image.

Taking the VisA dataset as an example, let $O=\{{\text{candle}, \text{cashew}, \dots \}}$ denote a collection of 12 object types.
%
If $o_i=\text{cashew}$, we select a normal cashew image as $x_{N}^{o_i}$. 
A designed question $Q_{o_i}$ can be regarded as ``What kind of anomaly may happen to cashew?".
By leveraging the powerful general knowledge, VLLMs (e.g. GPT-4V) may respond $A_{o_i}=\{\text{cracked, moldy...}\}$. 
Take $d=\text{cracked}$ as an example, the embedding of text ``cracked" will be used to match the most similar texture image in Tex-9K.

\subsection{Adaptive Texture Anomaly Generation}
\label{sec.4.4}
After obtaining the matched texture image, we propose Adaptive Texture Anomaly Generation, which seamlessly integrates the matched texture into a normal industrial image, creating a realistic anomaly sample, detailed in Alg~\ref{alg2}.

Specifically, we first generate an adaptive texture image $x_{\wp}$ based on the matched texture image and the normal industrial object image $x_N$ (dropping the superscript for simplicity), along with an inpainting mask $M_{in}$ that indicates the region where the anomaly content is to be generated. We discuss how to get $M_{in}$ and $x_{\wp}$ later in \textbf{Mask Generation}.

\begin{figure}[!t]
  \centering
  \includegraphics[width=1\linewidth]{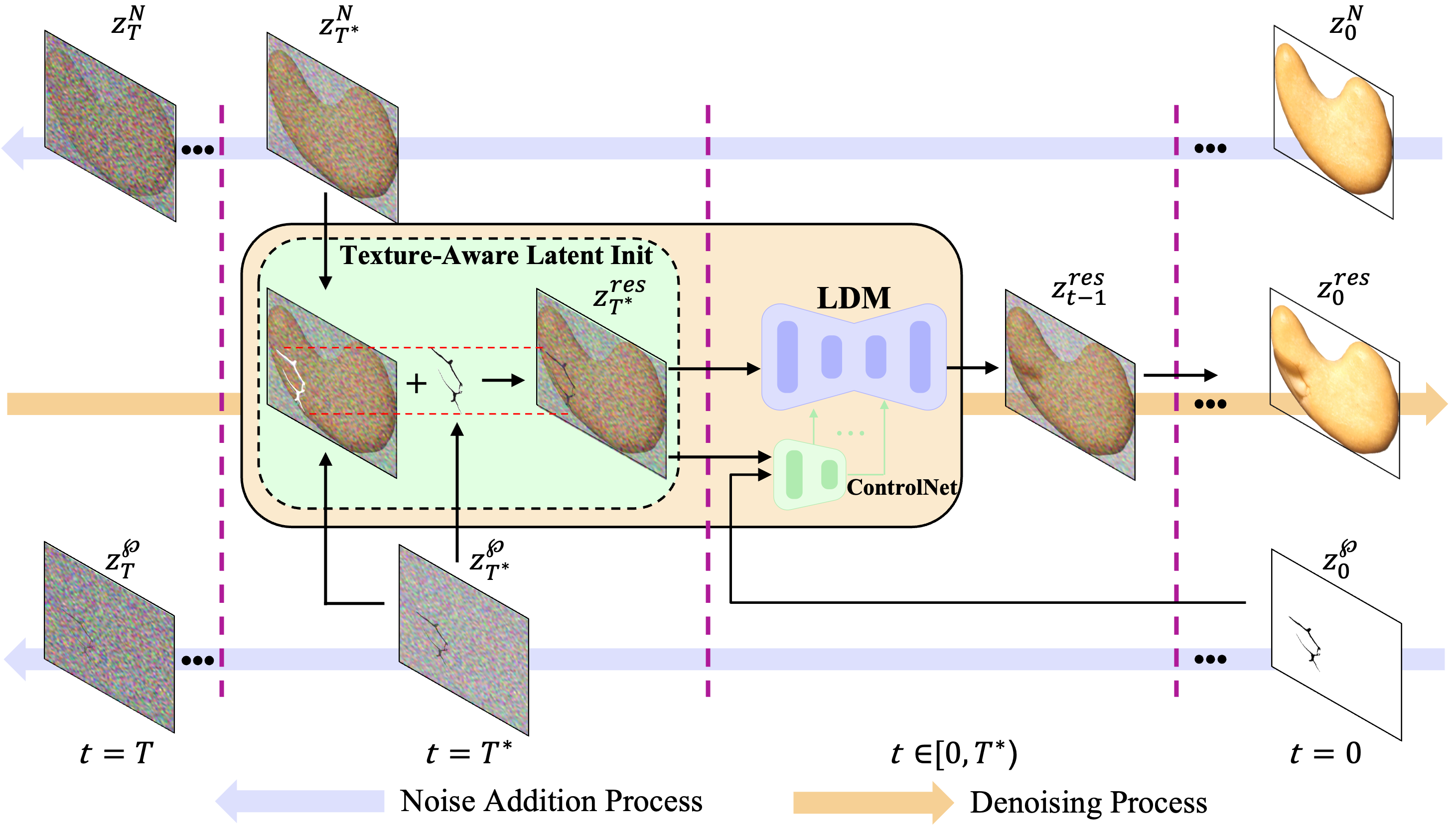}
  \caption{
    Texture-Aware Latent Initialization (TALI) blends normal image latent $z^{N}$ and adaptive texture latent $z^{\wp}$ at a later timestep $T^*$ to get $z_{T^*}^{\text{res}}$ as the starting point for better denoising result. For better clarity, the images are shown in the pixel space instead of the latent space.
  }
  \label{fig.6}
\end{figure}

To realistically express texture variations in the normal image $x_N$ in region $M_{in}$, we pioneer the application of  ControlNet, which offers powerful edge-mask control ability. 
Since ControlNet is trained on natural images, it often becomes unstable on industrial data, leading to subtle or inconsistent defect generation. To address this, we introduce Texture-Aware Latent Initialization (TALI), enhancing its reliability in industrial anomaly synthesis (Figure~\ref{fig.6}).

\noindent \textbf{Texture-aware Latent Init}. 
The core idea of TALI is to blend the normal industrial image $x_{N}$ with the adaptive texture image $x_{\wp}$ in the latent space as the starting point for denoising, in combination with ControlNet to achieve more stable anomaly generation. 
In fact, this part can be regarded as an image composition task, which aims to harmonize $x_{N}$ and $x_{\wp}$ to generate the composite anomaly image $x_{res}$. 
Unlike traditional image composition methods, such as TF-ICON~\cite{lu2023tf}, which typically invert $x_{N}$ and $x_{\wp}$ into their corresponding noisy latent representations $z_{T}^{N}$ and $z_{T}^{\wp}$ at a predefined timestep $T$, we choose to begin at a later timestep $T^*$. 
This choice helps avoid excessive stylistic disconnection between the generated anomalous part of $x_{res}$ and the normal industrial image $x_{N}$, while ensuring anomalous part of $x_{\text{res}}$ remains geometrically stable and consistent with $x_{\wp}$. 
Mathematically, we choose $0 < T^* < T$ and use $z_{T^*}^{\text{res}}$ as the starting point for denoising:
\begin{equation}
\mathbf{z}_{T^{*}}^{res}=\mathbf{z}_{T^{*}}^{N}\odot(\mathbf{1}-\mathbf{M}_{\wp}^z)+\mathbf{z}_{T^{*}}^{\wp}\odot\mathbf{M}_{\wp}^z,
\end{equation}
where $\mathbf{M}_{\wp}^z$ is the segmentation mask of $x_{\wp}$ in latent space.

After initialization, the pre-trained denoising network  $\epsilon_{\theta}$ with ControlNet will preserve the layout structure of  $z_{T^{*}}^{res}$  during  $t \in [0, T^{*}]$, while gradually harmonizing the anomalous texture with the normal image in the inpainting mask $M_{in}$ region. 
The denoising process is achieved through the removal of the estimated noise $\epsilon_{\theta}(z_t^{res},t, M_{in}, x_{\wp},P)$, where $M_{in}$ defines the region of the original image that can be inpainted, $x_{\wp}$ serves as the ControlNet condition, $P$ is an object-specific text prompt embedding. 
The text prompt is formulated as: “A photo of [$d$] [$o_i$]”, where, for example, $d$ could be “cracked” generated by VLLM, and $o_i$ refers to an industrial category such as “cashew”.
When the denoising reaches $t = 0$, the decoder $\mathcal{D}$ is used to decode and obtain $x^{res} = \mathcal{D}(z_{0}^{res})$. We detail $T^{*}$ selection in Sec~\ref{ablation}.

\begin{figure}[!t]
  \centering
  \includegraphics[width=1\linewidth]{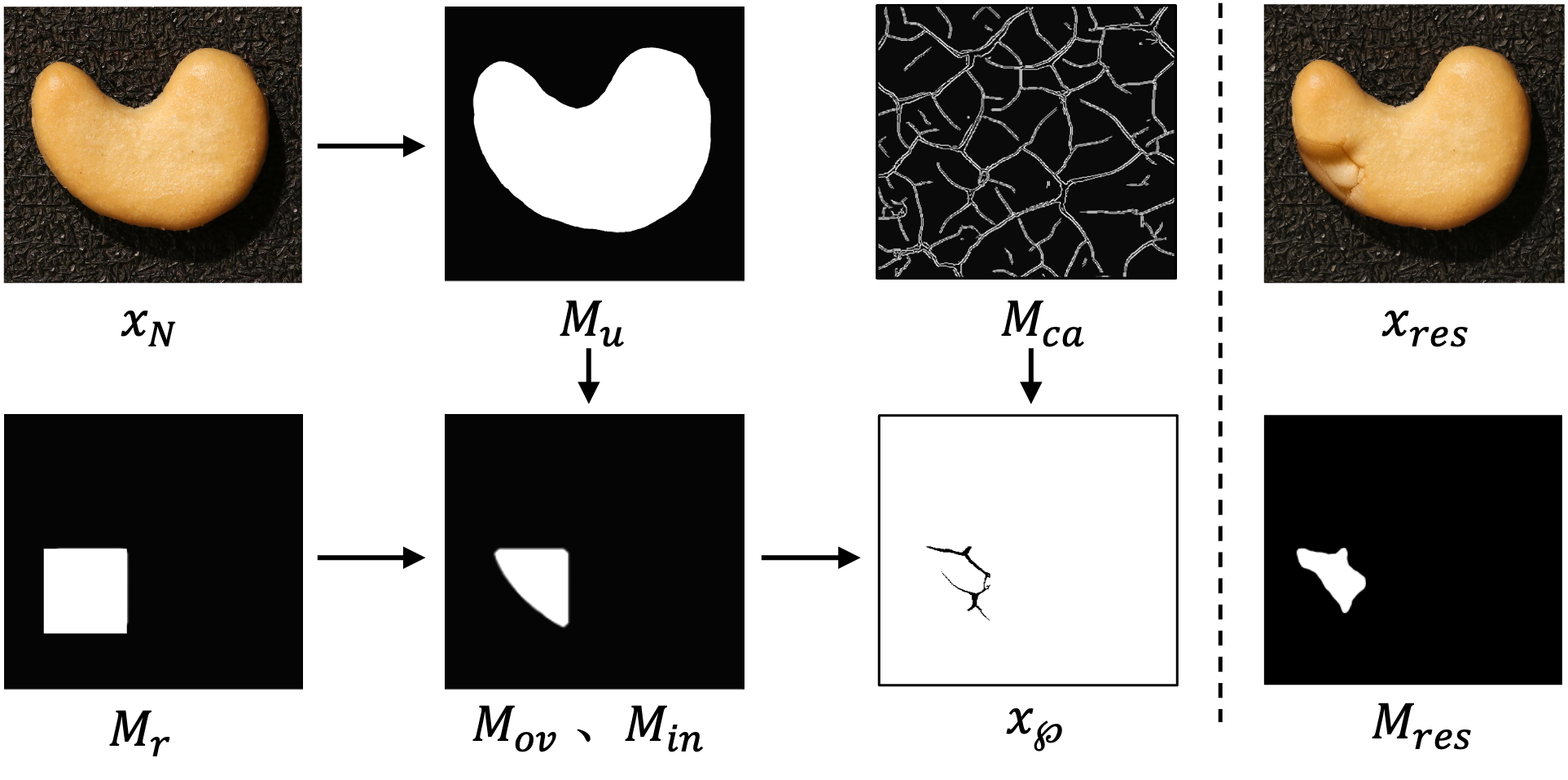}
  \caption{
  \textbf{Left}: An example of a successfully generated inpainting mask $M_{\text{in}}$ and its corresponding adaptive texture image $x_{\wp}$. \textbf{Right}: An example of the generated anomaly result and the refined mask.
  }
  \vspace{-10pt}
  \label{fig.7}
\end{figure}

\noindent \textbf{Mask Generation}. 
The previous anomaly synthesis methods rely on arbitrary inpainting masks, such as Perlin noise~\cite{zhang2024realnet} or random masks~\cite{hu2024anomalydiffusion}, which often result in anomaly being placed on the background of industrial objects, reducing realism. 
To address this, we propose the following strategy for generating a more effective inpainting mask: 
(a) Randomly generate a rectangular mask $M_r$. 
(b) Compute the intersection $M_\text{ov}=M_r \odot M_u$ between $M_r$ and foreground $M_u$ (object segmentation~\cite{Qin_2020_PR} in normal image).
Proceed only if the overlap $Area(M_\text{r} \odot M_u) >\text{thresh}_1$. 
(c) Compute the intersection between $M_\text{ov}$ and $M_{\text{ca}}$ (Canny edges~\cite{canny1986computational} of matched texture). 
If the overlap $Area(M_{ov}\odot M_{ca})>\text{thresh}_2$, we accept $M_\text{ov}$ as the inpainting mask $M_{\text{in}}$, and use the corresponding texture region as the anomaly texture $x_{\wp}$. The final inpainting mask is obtained as $M_{\text{in}} = M_r \odot M_u$, and adaptive texture $x_{\wp}$ is generated by applying morphological operations to $M_{\text{in}} \odot M_{ca}$ to ensure connectivity, detailed in \textit{Appendix.C}.
%
This strategy generates a random and diverse mask while ensuring valid texture. We present an example of a generated mask and its corresponding adaptive texture image in Figure~\ref{fig.7} (left).

\noindent \textbf{Mask Refine}. Since the inpainting mask $M_{\text{in}}$ covers a large area, the LDMs’ generation process is relatively unconstrained, often resulting in anomalies appearing only in certain regions.
A coarse refinement is then applied by computing the SSIM~\cite{wang2004ssim} difference map between the original image $x_N$ and the generated anomaly image $x_{res}$ within the inpainting region $M_{in}$, yielding a per-pixel similarity score, i.e. $\text{ScoreMap}_{\text{ssim}}=\text{SSIM}(x_N \odot M_{\text{in}}, x_{\text{res}} \odot M_{\text{in}})$. We then compute the mean similarity $\text{thresh}_{\text{mean}}=\text{Average}(\text{ScoreMap}_{\text{ssim}})$ within the region and retain pixels with similarity scores exceeding this threshold, i.e. $M_{\text{res}} = \text{ScoreMap}_{\text{ssim}}>\text{thresh}_{\text{mean}}$. 
We show an example of generated anomaly image with refined mask in Figure~\ref{fig.7} (right).

\begin{algorithm}[h]

\caption{Anomaly Description Generation \& Matching}

 \hspace*{0.02in} \textbf{Data:}  \( O = \{o_1, o_2, \dots \} \), \( \mathcal{X} = \{x_{\text{tex}_1}, x_{\text{tex}_2}, \dots, x_{\text{tex}_m} \} \)
 \hspace*{0.02in} \textbf{Cache:} $e_{\mathcal{X}}=\{e_1,e_2, \dots, e_m\} \gets \text{CLIP}_{\text{image}}(\mathcal{X})$
 
 \hspace*{0.02in} \textbf{Input:} \( o_i \in O \), a normal image \( x_N^{o_i} \)
 
 \hspace*{0.02in}  \textbf{Output:} Matched descriptions and texture images for \(o_i\)
 
 \begin{algorithmic}[1]
    \State Pose question \( Q_{o_i} = \text{Template}(o_i) \) 
    \State  Obtain anomaly descriptions \(  A_{o_i} =  \text{VLLM}(  Q_{o_i} , x_N^{o_i}  )\)  
    \For {description  \( d  \text{  in  } A_{o_i}= \{d_1, d_2, \dots, d_k \}\)}
        \State Obtain description embedding \( e_{d} \gets \text{CLIP}_\text{text}(d) \)
    
        \State Obtain matched embedding
        \vspace{-4mm} 

        \[        e_{\text{match}} \gets \arg \max_{e_i \in e_{\mathcal{X}}} \space \cos \langle e_{d}, e_i \rangle\]
        \vspace{-4mm} 
        \State Select the corresponding texture image \( x_{\text{match}} \in \mathcal{X} \)
    \EndFor
    \State \Return All descriptions and matched texture images
\end{algorithmic}
\label{alg1}
\end{algorithm}

\begin{algorithm}[!h]
\caption{Adaptive Texture Anomaly Generation}
\label{alg2}
\hspace*{0.02in} \textbf{Input:}  $x_N$, $x_\text{match}$, $T^*$, P.

\hspace*{0.02in} \textbf{Cache:} $M_u \gets \text{Seg}(x_N)$,$M_\text{ca} \gets \text{Canny}(x_\text{match})$

\hspace*{0.02in} \textbf{Output:}  Generated anomaly image $x_\text{res}$ and mask $M_\text{res}$.
\begin{algorithmic}[1]
    
\State $M_{in}, x_{\wp}$ = \textbf{Mask Generation}($M_u,M_\text{ca}$)
\State \textbf{TALI:}
\State $z_N,z_{\wp} =\mathcal{E}(x_N),\mathcal{E}(x_{\wp}) $

\State Sample Noise $\epsilon \sim \mathcal{N}(0,1)$
\State Add Noise $z_{T^*}^{N} \gets  \sqrt{\bar{\alpha}_{T^*}} \space z_N + \sqrt{1 - \bar{\alpha}_{T^*}} \space \epsilon$
\State Add Noise $z_{T^*}^{\wp} \gets  \sqrt{\bar{\alpha}_{T^*}} \space z_\wp + \sqrt{1 - \bar{\alpha}_{T^*}} \space \epsilon$
\State Blend latents $z_{T^*}^\text{res} \gets z_{T^*}^{N}\odot (1 - M_\wp^z) + z_{T^*}^{\wp}\odot M_\wp^z$
\State \textbf{Denoise with ControlNet:}
\For {$t \gets T^*$ \textbf{downto} $1$}
    \State $z_{t-1}^{res} \gets \text{DDIM}\big(z_t^\text{res},\epsilon_{\theta}(z_t^\text{res}, t, M_\text{in}, x_{\wp}, P)\big)$.
\EndFor
\State $x_\text{res} \gets \mathcal{D}(z_0^\text{res})$.
\State $M_\text{res}$ = \textbf{Mask Refine}($x_\text{N},x_\text{res},M_\text{in}$)
\State \Return $x_\text{res},M_\text{res}$.

\end{algorithmic}
\end{algorithm}

\begin{figure*}[!t]
  \centering
  \includegraphics[width=1\linewidth]{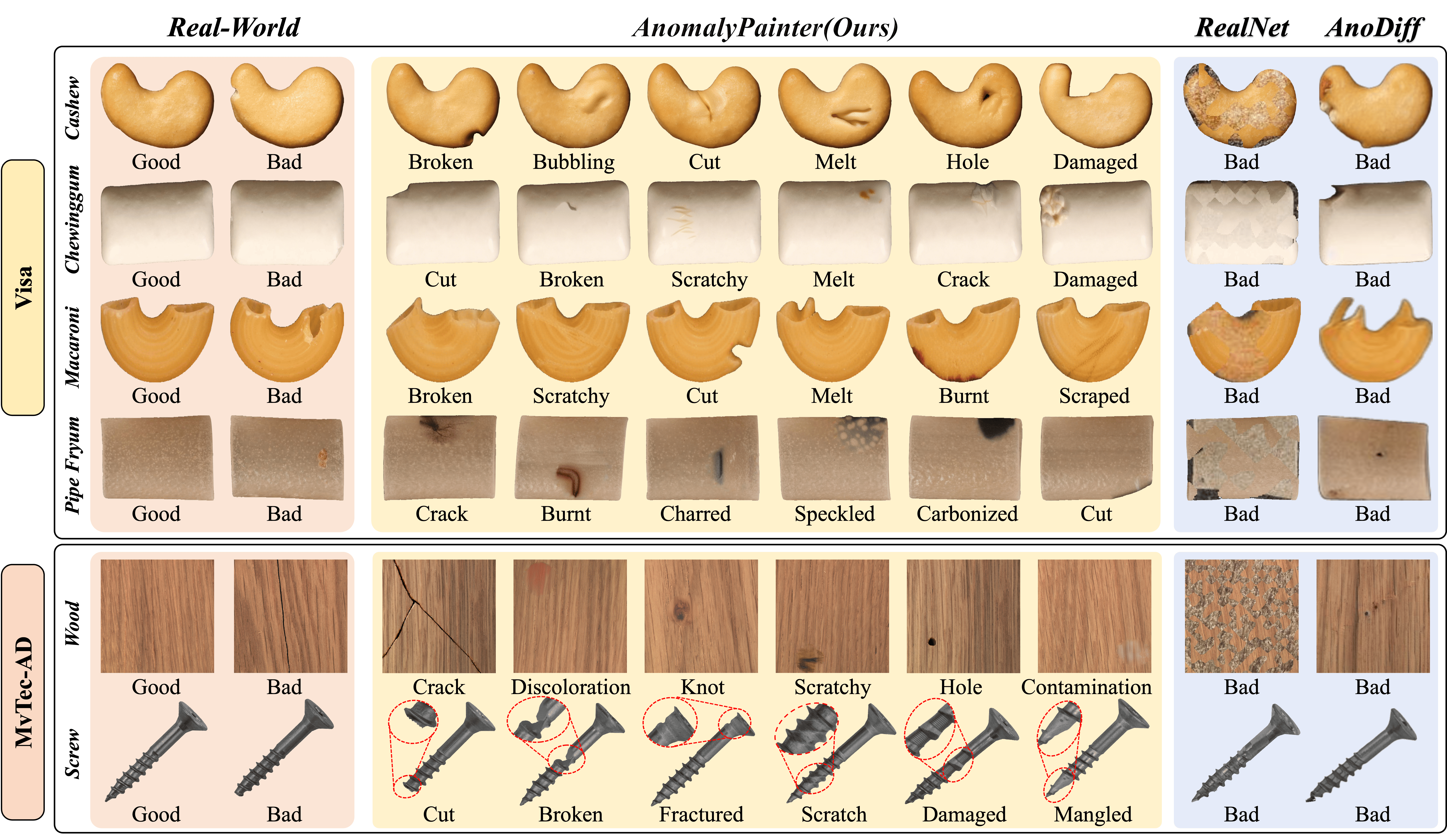}
  \caption{
  Qualitative Comparison. It is clear that our method can generate diverse and realistic anomaly data across various industrial objects in multiple datasets. It outperforms both the best few-shot method, AnoDiff, and the best zero-shot method, RealNet.
  }
  \label{visual-main}
  \vspace{-5pt}
\end{figure*}

\section{Experiments}
\subsection{Experimental Settings}
\noindent \textbf{Dataset}.~We conduct experiments on the MVTec AD~\cite{bergmann2019mvtec} and VisA~\cite{zou2022spot} datasets. MVTec AD consists of 15 categories with 5,354 images, suitable for single-object/texture anomaly inspection. VisA contains 12 categories with 10,821 images, featuring complex objects and multi-object categories. Unlike the fine-grained classification of MVTec AD, VisA employs a weak-category prior evaluation paradigm, which is more aligned with the needs of detecting unknown anomalies in industrial settings and better reflects anomaly synthesis generalizability and robustness. Experiments are primarily evaluated on VisA, with additional results on other datasets provided in \textit{Appendix.D}.

\noindent \textbf{Evaluation Metrics}. 
For anomaly synthesis, we use Inception Score (IS) and intra-cluster pairwise LPIPS distance (IL) to assess synthesis quality and diversity. 
For anomaly inspection, we employ pixel-level and image-level AUROC as evaluation metrics.

\noindent \textbf{Implementation Details}. 
\modelname is implemented using the HuggingFace Diffusers library~\cite{diffusers}, built on the Stable Diffusion XL 1.0 (SDXL)~\cite{podell2024sdxl} model and ControlNet-Canny~\cite{controlnet_canny}, with a CLIP model utilizing a ViT-B/32 backbone.
We adopt GPT-4V as our VLLM. Following~\cite{lu2023tf}, we use a 20-step DDIM sampler, while starting denoising at $T^*=16$.
Similar to~\cite{zavrtanik2021draem,hu2024anomalydiffusion}, we generate 500 images per anomaly category and train a U-Net~\cite{ronneberger2015u} model for downstream anomaly inspection tasks. 
As our method is training-free, each anomaly image synthesis takes only 6 seconds on a single Nvidia GeForce RTX 3090 GPU.

\noindent \textbf{Baselines}. 
We select a variety of high-performing open-sourced methods as baselines for downstream tasks, including zero-shot methods that include Cut-Paste~\cite{li2021cutpaste}, DRAEM~\cite{zavrtanik2021draem}, and RealNet~\cite{zhang2024realnet}, and few-shot methods that include DFMGAN~\cite{duan2023few}, AnoGen~\cite{gui2024anogen}, and AnoDiff~\cite{hu2024anomalydiffusion}. For anomaly synthesis comparison, we choose the best-performing zero-shot and few-shot methods, RealNet and AnoDiff, for visualization and quality comparison.

\begin{table*}[ht]
\centering
\setlength{\tabcolsep}{11pt} 
\renewcommand{\arraystretch}{0.89} 
\resizebox{\textwidth}{!}{ 
\begin{tabular}{c|ccc|cccc}
\toprule
\multirow{2}{*}{Category} & \multicolumn{3}{c|}{Few-Shot}     & \multicolumn{4}{c}{Zero-Shot}                 \\ \cmidrule(lr){2-8}
                          & DFMGAN    & AnoGen    & AnoDiff   & CutPaste  & DRAEM     & RealNet   & Ours      \\ \midrule
\textit{candle}                    & 63.9/81.2 & 83.7/88.6 & 87.9/91.2 & 81.9/89.9 & 83.1/91.6 & 88.9/93.8 & \textbf{98.7}/\textbf{98.4} \\
\textit{capsules}                  & 69.1/69.4 & 76.7/75.6 & 79.6/80.1 & 91.8/90.8 & 92.5/92.5 & \textbf{95.2}/92.3 & 93.3/\textbf{94.6} \\
\textit{cashew}                    & 93.2/94.7 & 95.1/95.3 & 95.2/95.9 & 97.2/94.7 & 98.2/96.3 & 98.6/96.3 & \textbf{98.9}/\textbf{98.4} \\
\textit{chewinggum}                & 95.3/99.1 & 96.4/98.8 & 97.3/98.9 & 95.5/96.9 & 96.6/98.0 & 97.8/98.9 & \textbf{99.1}/\textbf{99.5} \\
\textit{fryum}                     & 91.9/96.1 & 93.5/96.0 & 93.1/95.6 & 79.9/89.3 & 80.2/89.2 & 86.1/92.5 & \textbf{90.7}/\textbf{95.0} \\
\textit{macaroni1}                 & 90.2/95.5 & 93.7/95.6 & \textbf{94.4}/95.2 & 84.7/90.3 & 82.3/90.0 & 89.8/93.1 & 93.4/\textbf{98.5} \\
\textit{macaroni2}                 & 49.8/91.8 & 73.7/89.7 & 68.5/91.4 & 86.3/82.8 & 84.0/81.9   & \textbf{91.7}/93.8 & 87.8/\textbf{96.6} \\
\textit{pcb1}                      & 93.8/96.3 & 94.3/94.7 & \textbf{95.8}/\textbf{96.3} & 92.9/92.3 & 94.5/93.5 & 93.9/94.2 & 93.3/95.4 \\
\textit{pcb2}                      & 83.9/\textbf{94.8} & 85.8/89.8 & 84.5/90.8 & 81.7/88.1 & 81.3/88.5 & 85.5/89.1 & \textbf{88.9}/93.9 \\
\textit{pcb3}                      & 80.6/83.5 & 81.6/83.1 & 84.3/85.1 & 57.6/78.3 & 69.4/78.7 & 71.3/83.2 & \textbf{89.2}/\textbf{92.5} \\
\textit{pcb4}                      & 94.3/91.5 & 94.4/90.6 & \textbf{96.2}/92.2 & 84.3/83.1 & 86.7/87.0 & 89.7/89.0 & 95.3/\textbf{93.5} \\
\textit{pipe fryum}               & 82.3/95.9 & 91.2/96.7 & 93.9/97.4 & 82.3/97.5 & 84.4/98.2 & 94.0/97.3 & \textbf{98.7}/\textbf{98.6} \\ \midrule
\textit{Average}                      & 82.4/90.8 & 88.3/91.2 & 89.2/92.5 & 84.7/89.5 & 86.1/90.4 & 90.2/92.8 & \textbf{93.9}/\textbf{96.2} \\ \bottomrule
\end{tabular}
} 
\vspace{-5pt}
\caption{Comparison of AUC-I/AUC-P across image-level and pixel-level anomaly localization on VisA dataset by training a U-Net on the synthesized data from DFMGAN, AnoGen, AnoDiff, CutPaste, DRAEM, RealNet, and our AnomalyPainter.}
\vspace{-5pt}
\label{tab:comparison}
\end{table*}

\begin{figure}[!h]
  \centering
  \includegraphics[width=1\linewidth]{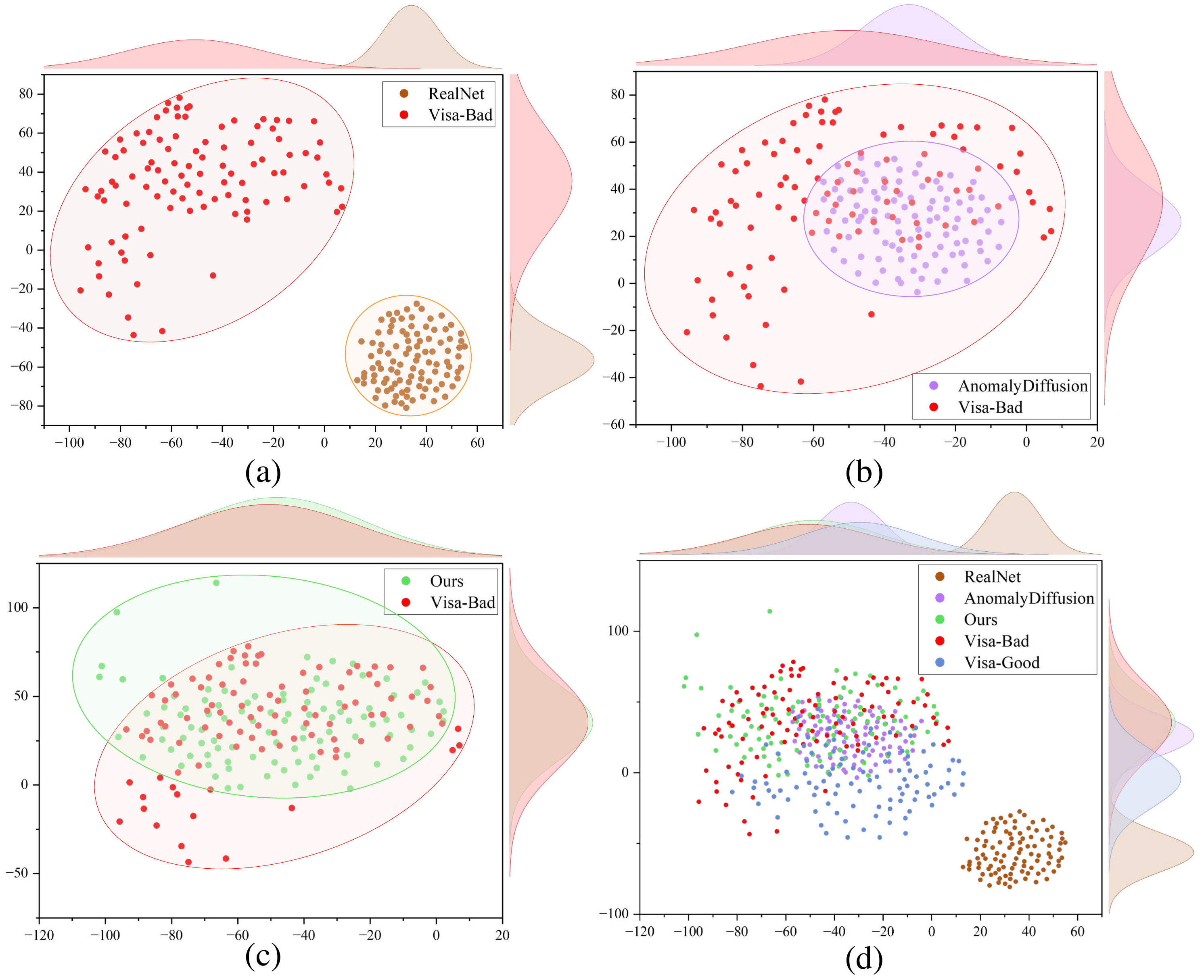}
  \caption{
  Marginal Group Plot of \textbf{cashew} (object in VisA) t-SNE results for anomaly samples synthesized by three methods, real anomaly samples (VisA-Bad) from test set, and normal samples (VisA-Good). 
  See \textit{Appendix.E} for more examples and details.
  }
  \label{fig.8}
  \vspace{-15pt}
\end{figure}
\subsection{Comparison in Anomaly Generation}
\textbf{Qualitative Comparison}. 
We present anomaly images synthesized by different methods on the VisA and MVTec datasets, as shown in Fig~\ref{visual-main}. Although RealNet can generate anomaly images with noticeable defects, the results often appear visually unrealistic and confusing. AnoDiff struggles with effective anomaly generation, particularly on challenging datasets like VisA, where it maps different anomaly features into the same embedding, leading to feature confusion, lack of diversity, and uniformity. Furthermore, due to architectural constraints, its generated images are blurry (256×256 resolution), with unnatural transitions between anomalies and objects. In contrast, our AnomalyPainter produces anomaly images with a resolution of up to 1024×1024, preserving structural details with natural transitions, higher realism, and greater diversity.

\begin{table}[!h]
    \centering
    \renewcommand{\arraystretch}{0.88} 
    \setlength{\tabcolsep}{7pt} 
    \resizebox{\linewidth}{!}{%
    \begin{tabular}{c|cc|cc|cc}
        \toprule
        \multirow{2}{*}{Category} & \multicolumn{2}{c|}{RealNet} & \multicolumn{2}{c|}{AnoDiff} & \multicolumn{2}{c}{Ours} \\
        \cmidrule(lr){2-7}
        & IS & IL & IS & IL & IS & IL \\
        \midrule
        \textit{candle}      & 1.33 & 0.27 & 1.33 & 0.18 & \textbf{1.65} & \textbf{0.29} \\
        \textit{capsules}    & 1.65 & \textbf{0.44} & 1.32 & 0.33 & \textbf{1.67} & 0.39 \\
        \textit{cashew}      & 1.65 & 0.31 & 1.29 & 0.27 & \textbf{1.83} & \textbf{0.36} \\
        \textit{chewinggum}  & 1.69 & 0.37 & 1.27 & 0.36 & \textbf{1.77} & \textbf{0.43} \\
        \textit{fryum}       & 1.35 & 0.22 & 1.13 & 0.18 & \textbf{1.69} & \textbf{0.28} \\
        \textit{macaroni1}   & \textbf{1.75} & \textbf{0.22} & 1.50 & 0.21 & 1.73 & 0.21 \\
        \textit{macaroni2}   & 1.78 & 0.32 & 1.61 & 0.24 & \textbf{1.92} & \textbf{0.41} \\
        \textit{pcb1}        & 1.47 & 0.33 & 1.22 & \textbf{0.35} & \textbf{1.49} & 0.34 \\
        \textit{pcb2}        & 1.37 & \textbf{0.33} & \textbf{1.56} & 0.30 & 1.53 & 0.31 \\
        \textit{pcb3}        & 1.26 & 0.19 & 1.21 & 0.23 & \textbf{1.51} & \textbf{0.26} \\
        \textit{pcb4}        & 1.35 & \textbf{0.30} & 1.27 & 0.28 & \textbf{1.50} & 0.27 \\
        \textit{pipe fryum}  & 1.53 & 0.22 & 1.34 & 0.22 & \textbf{1.69} & \textbf{0.36} \\
        \midrule
        \textit{Average}        & 1.52 & 0.29 & 1.34 & 0.26 & \textbf{1.67} & \textbf{0.33} \\
        \bottomrule
    \end{tabular}
    }
    \vspace{-5pt}
    \caption{Quantitative comparison on IS and IL on VisA.}
    \vspace{-15pt}
    \label{tab:ano_real_ours}
\end{table}

\noindent \textbf{Quantitative Comparison}. 
On the VisA dataset, we synthesize 500 anomaly images per object category and compute IS and IL metrics (Table~\ref{tab:ano_real_ours}). The results demonstrate that AnomalyPainter outperforms other methods in both diversity and realism.

\noindent \textbf{Data Distribution Comparison}. 
We use t-SNE~\cite{van2008visualizing} to visualize anomaly samples synthesized by three methods (RealNet, AnoDiff, and ours), along with real anomalies (VisA-Bad) and normal samples (VisA-Good) from the test set (Fig.~\ref{fig.8}). The overall results in Fig.~\ref{fig.8} (d) show that the distribution of anomalies synthesized by AnomalyPainter closely aligns with VisA-Bad, while AnoDiff’s anomalies are restricted to a limited region within VisA-Bad, and RealNet’s anomalies deviate further from VisA-Bad. Additionally, VisA-Bad samples are located near VisA-Good. To provide a clearer comparison of the distribution of synthesized anomalies relative to VisA-Bad, we present the distribution of each method separately in Fig.~\ref{fig.8} (a–c). The marginal distributions along the axes and the overlap of distribution ellipses in Fig.~\ref{fig.8} further validate the theoretical assumptions proposed in Fig~\ref{fig:teaser}.

\subsection{Comparison in Downstream Tasks}
\begin{figure}[t]
  \centering
  \includegraphics[width=1\linewidth]{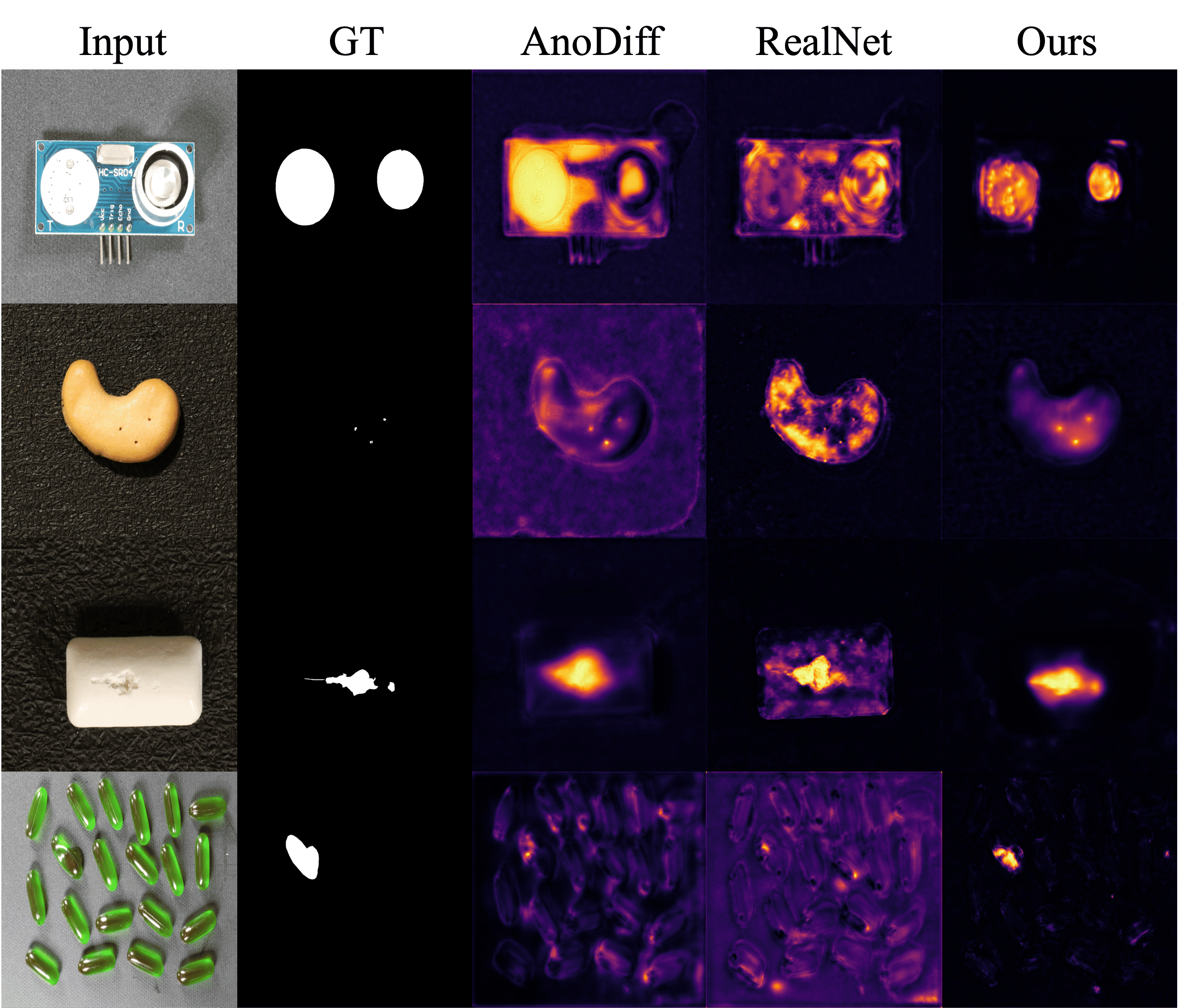}
  \caption{
  Quantitative anomaly localization comparison with a U-Net trained on the data synthesized by three methods. It shows that ours achieves the best anomaly localization results.
  }
  \label{heatmap}
\end{figure}

\textbf{Anomaly Synthesis for Detection and Localization.}
We compare our method with existing anomaly generation approaches to evaluate its effectiveness in downstream anomaly detection and localization tasks. For each object category, we compute image-level and pixel-level AUROC, AP, and F1-Max scores (due to table width limitations, only AUROC is presented, while AP and F1-Max results are provided in \textit{Appendix D}). As shown in Table~\ref{tab:comparison}, the U-Net~\cite{ronneberger2015u} trained on data generated by our method achieves the highest AUC-I (\textbf{93.9\%}) and AUC-P (\textbf{96.2\%}) on the VisA dataset, outperforming the SOTA zero-shot method RealNet by \textbf{3.7\%} and \textbf{3.4\%}, and the SOTA few-shot method AnoDiff by \textbf{4.7\%} and \textbf{3.7\%}, respectively. Additionally, Figure~\ref{heatmap} presents qualitative comparisons of anomaly localization, further highlighting our approach.

\begin{table}[t]
\centering
\renewcommand{\arraystretch}{0.9} 
\resizebox{0.48\textwidth}{!} 
{
\begin{tabular}{@{}ccc|cccc@{}}
\toprule
 ADGM & TALI & ControlNet &  IS   & IL & AUC-I & AUC-P \\ \midrule
\checkmark   & -  & -    & 1.58 & 0.25 & 90.5 & 92.7  \\
\checkmark   & \checkmark  & -    & 1.62 & 0.29 & 92.7 & 94.2  \\
\checkmark    & -  & \checkmark  & 1.63 & 0.28 & 92.3  & 94.7  \\
- & \checkmark  & \checkmark   & 1.53 & 0.26 & 92.1  & 93.5  \\
\checkmark    & \checkmark   & \checkmark         & \textbf{1.67} & \textbf{0.33} & \textbf{93.9}  & \textbf{96.2}
  \\ \bottomrule
\end{tabular}
}
\caption{Ablation in VisA on our Anomaly Description Generation and Matching (ADGM), Texture-Aware Latent Initialization (TALI) and ControlNet in Adaptive Anomaly Texture Generation. The results indicate that omitting any of the proposed components leads to a noticeable decline in performance.}
\label{ablation1}
\vspace{-15pt}
\end{table}

\subsection{Ablation Study}
\label{ablation}
We evaluate the effectiveness of our components: Anomaly Description Generation and Matching (ADGM), Texture-Aware Latent Initialization (TALI) and ControlNet in Adaptive Anomaly Texture Generation. 
We design 5 different combinations to demonstrate the effectiveness of each component in Table.~\ref{ablation1}. 
First, with only ADGM module, our method degrades to a crop-and-paste method, resulting in a significant decrease in realism with  $\text{IL}=0.25$. 
Next, we introduce TALI and ControlNet separately, both of which improve the realism, with IL increasing to 0.29 and 0.28. 
When TALI and ControlNet are combined, the realism improves further, with IL reaching 0.33, which also performs best in downstream detection tasks.
We also run the full TALI and ControlNet experiment without the ADGM module, using random textures as guidance. 
As expected, random textures often lead to mismatched anomalies, reducing realism with IL decreasing to 0.26.

\begin{table}[t]
\resizebox{0.48\textwidth}{!}{
\begin{tabular}{@{}c|c|c|c|c|c@{}}
\toprule
Choice      & $T^*$=20     & $T^*$=18     & $T^*$=16              & $T^*$=14     & $T^*$=12     \\ \midrule
IS/IL       & 1.64 / 0.30 & \textbf{1.69} / 0.32 & 1.67 / \textbf{ 0.33} & 1.65 / 0.31 & 1.59 / 0.32 \\
AUC-I/AUC-P & 92.8 / 94.9 & 93.4 / 95.4 & \textbf{93.9 / 96.2} & 93.1 / 95.1 & 92.9 / 95.1 \\ \bottomrule
\end{tabular}
}
\vspace{-8pt}
\caption{Quantitative Ablation with different $T*$.}
\label{tstar}
\vspace{-8pt}
\end{table}

\begin{figure}[t]
  \centering
  \includegraphics[width=1\linewidth]{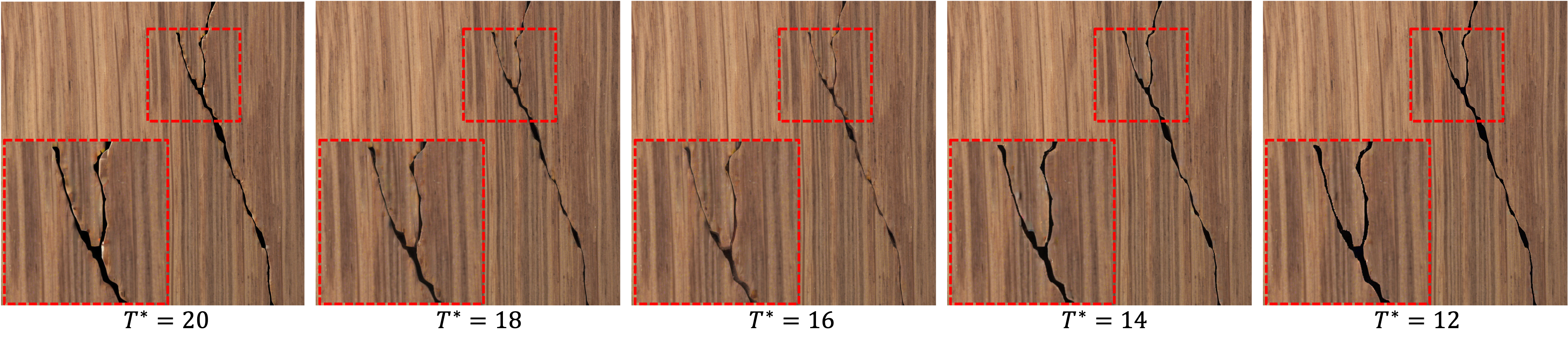}
  \vspace{-18pt}
  \caption{
    "A photo of cracked wood" results with differet $T^*$.
  }
   \vspace{-15pt}
  \label{wood_ablation}
\end{figure}

\noindent \textbf{$T^*$ Selection}.
Following TF-ICON~\cite{lu2023tf}, we adopt the common practice of setting  $T = 20$.
To find best $T^*$ later than $T^*$, we experiment with the influence of different choices of $T^*$ and give quantitative results in Table.~\ref{tstar}. 
A typical qualitative visualization on wood is also shown in Figure~\ref{wood_ablation}.
It is clear that too large step (e.g. $T^*=20$) may cause the inpainting intensity to be too strong, while too small step (e.g. $T^*=12$) can result in an overly strong initialization binding, making the anomaly content appear excessively unnatural, thus decreasing the realism.
\section{Conclusion}
We propose \modelname, a novel zero-shot anomaly synthesis method that synthesizes diverse and realistic anomaly data.
\modelname combines the ability of VLLMs and LDMs with our proposed Tex-9K to synthesize in a more real-world manner. 
Tex-9K consists of 75 categories and 8,792 texture images, designed to better support professional assets for the entire anomaly synthesis community.
Extensive experiments show that \modelname outperforms existing anomaly synthesis methods and our synthesized anomaly data effectively improves the performance of the downstream inspection tasks.

\noindent \textbf{Limitations}. Similar to previous methods, our method still struggles to synthesize global layout anomalies (e.g., the same object appearing multiple times in an image or objects swapping positions).

\noindent \textbf{Broader Impact}. This study pioneers a universal open-world anomaly synthesis approach, enabling unknown anomalies synthesis without prior industrial anomaly samples. This breakthrough lays a crucial foundation for both industry and academia, paving the way for more adaptive and scalable anomaly detection systems.
\clearpage
{
    \small
    \bibliographystyle{ieeenat_fullname}
    \bibliography{main}
}
\clearpage
\clearpage
\setcounter{page}{1}

\appendix

\section{Tex-9K Construction}
\begin{figure*}[p]
  \centering
  \includegraphics[width=0.9\linewidth]{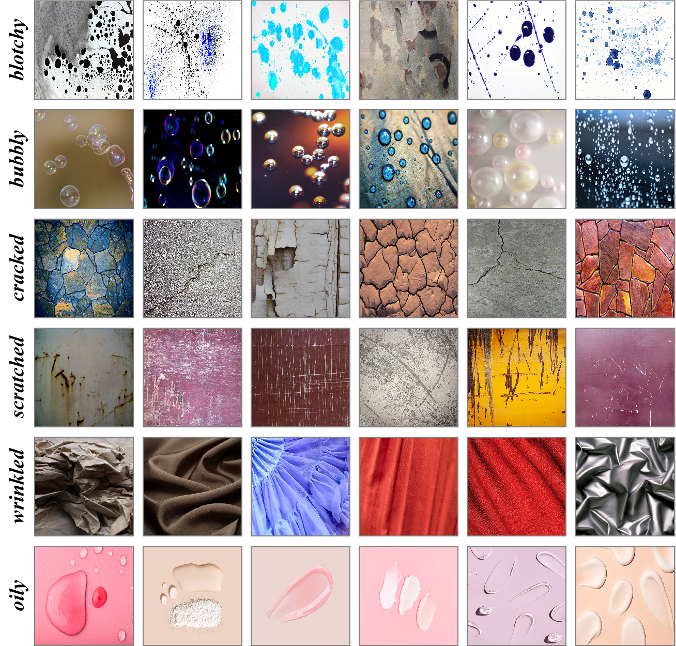}
  \caption{
    Examples of our proposed Tex-9K.
  }
  \label{fig.10}
\end{figure*}

We construct Tex-9K to provide a diverse and high-quality selection of texture patterns for fine-grained anomaly synthesis. Specifically, for 75 selected anomaly categories, we use web crawlers to collect 200 relevant images per category from Google. Additionally, if suitable images are available in existing datasets (e.g., DTD~\cite{cimpoi14describing}), we integrate them to enhance coverage.
To ensure texture clarity, we apply an automated density filtering process. Using the Canny edge detector~\cite{canny1986computational}, we generate edge masks for all images and discard those where edge coverage exceeds 70\% or falls below 2\% of the total image area. The remaining images undergo careful manual selection, resulting in a curated dataset of 8,792 high-quality texture images. Figure~\ref{fig.10} provides examples from Tex-9K.

\begin{figure*}[p]
  \centering
  \includegraphics[width=0.9\linewidth]{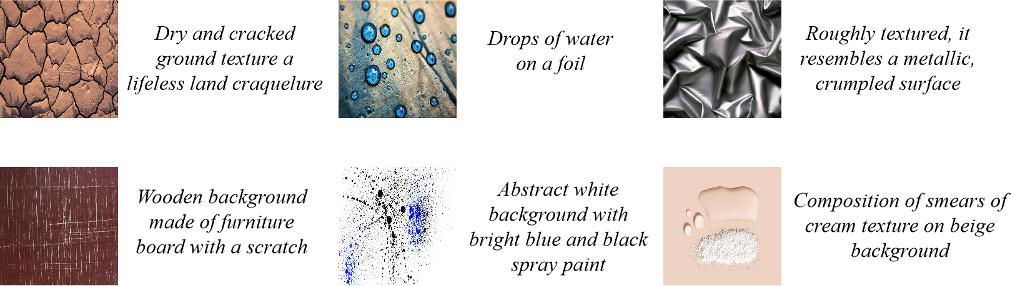}
  \caption{
    Examples from Tex-9K with corresponding descriptive text prompts
  }
  \label{fig.11}
\end{figure*}

Furthermore, for each image, we generate descriptive text prompts using the multi-modal large language model InternLM-XComposer~\cite{dong2025internlm}, enhancing its usability for conditioned anomaly synthesis. See Figure~\ref{fig.11} for some examples. 

\section{VLLM question template}
For each industrial object category $o_i$, we use the template to query the GPT-4V~\cite{OpenAI2023GPT4V} API to obtain the answer:
\textit{``You are a professional industrial anomaly engineer. Now I am doing some tasks on industrial defect photo generation. Please carefully analyze the material of the object [$o_i$] in the image and provide possible defects that could occur on it. I don't need that much explanation, just give me the brief answer like "cracked, faded". Please make a specific analysis according to the specific situation."}
In fact, you can create a custom prompt based on your own needs to get the description you want.

\section{Mask Strategy}
\textbf{Mask Generation}

(1)Random Rectangle Mask Generation:
We randomly generate a rectangular mask on an $H \times W$ image, where the mask’s length is chosen from $[l_{\text{rate}} \times H, h_{\text{rate}} \times H]$ and the width from $[l_{\text{rate}} \times W, h_{\text{rate}} \times W]$. $l_{\text{rate}}$ and $h_{\text{rate}}$ are adjustable parameters. This results in a random mask $M_r$.

(2)Foreground Segmentation and Overlap Filtering:
A segmentation model extracts the foreground mask $M_u$ from normal object image $x_N$. We compute the intersection $M_\text{ov}$ between $M_r$ and $M_u$. If the overlap exceeds a threshold $\text{thresh}_1$, we proceed. Otherwise, we regenerate $M_r$ in (1).

(3)Texture Validity Check:
We compute the intersection of $M_\text{ov}$ and the Canny edge map $M_{ca}$ extracted from the matched texture image. If the valid overlap exceeds a threshold $\text{thresh}_2$, we accept $M_\text{ov}$ as the inpainting mask $M_{\text{in}}$ and adaptive texture $x_{\wp}$ is generated by applying morphological operations to $M_{\text{in}} \odot M_{ca}$ to ensure connectivity. If not, we regenerate $M_r$ in (1).

In the specific implementation, $l_{\text{rate}}$ and $h_{\text{rate}}$ are set to 0.1 and 0.3, respectively. $\text{thresh}_1$ s set to 0.3 and $\text{thresh}_2$  is set to 0.05. The morphological operations first invert the pixel values of $M_{\text{in}} \odot M_{ca}$, followed by a single 5×5 kernel dilation and subsequent erosion to get the complete and connected texture.


\section{More Comparisons and Downstream Experiments on Visa and Mvtec}

\noindent\textbf{More Quantitative Comparison and Analysis.} We evaluated various zero-shot and few-shot anomaly synthesis methods, including DFMGAN~\cite{duan2023few}, AnoGen~\cite{hu2024anomalydiffusion}, AnoDiff~\cite{hu2024anomalydiffusion}, CutPaste~\cite{li2021cutpaste}, DRAEM~\cite{zavrtanik2021draem}, RealNet~\cite{zhang2024realnet}, and our proposed AnomalyPainter. For each object category in the MVTEC~\cite{bergmann2019mvtec} dataset, we synthesized 500 anomaly images and computed the Inception Score (IS) and Inception Loss (IL) for all methods, with detailed results shown in Table~\ref{tab:all_ISIL_mvtec}. The experiments demonstrate that AnomalyPainter significantly outperforms the compared methods in terms of both diversity and realism of generated images. Additionally, supplementary experimental results on the Visa~\cite{zou2022spot} dataset, as presented in Table~\ref{tab:all_ISIL_visa}, further validate the superior performance of AnomalyPainter.
\begin{table}[h]
\renewcommand{\arraystretch}{1.0} 
    \setlength{\tabcolsep}{2pt} 
    \resizebox{\linewidth}{!}{%
\begin{tabular}{c|ccccccc}
\toprule
\multicolumn{1}{c|}{Metric} & \multicolumn{1}{c}{DFMGAN} & \multicolumn{1}{c}{AnoGen} & \multicolumn{1}{c}{AnoDiff} & \multicolumn{1}{c}{Cutpaste} & \multicolumn{1}{c}{DRAEM} & \multicolumn{1}{c}{RealNet} & \multicolumn{1}{c}{Ours} \\
\midrule
IS                         & 1.72                       & 1.68                       & 1.80                        & 1.52                         & 1.73                      & 1.70                        & \textbf{1.91}            \\
IL                         & 0.20                       & 0.23                       & 0.32                        & 0.15                         & 0.24                      & 0.22                        & \textbf{0.35}     \\
\bottomrule
\end{tabular}
}
\caption{Quantitative comparison of IS and IL on MVTec AD.}
\label{tab:all_ISIL_mvtec}
\end{table}
\begin{table}[h]
\renewcommand{\arraystretch}{1.0} 
    \setlength{\tabcolsep}{2pt} 
    \resizebox{\linewidth}{!}{%
\begin{tabular}{c|ccccccc}
\toprule
\multicolumn{1}{c|}{Metric} & \multicolumn{1}{c}{DFMGAN} & \multicolumn{1}{c}{AnoGen} & \multicolumn{1}{c}{AnoDiff} & \multicolumn{1}{c}{Cutpaste} & \multicolumn{1}{c}{DRAEM} & \multicolumn{1}{c}{RealNet} & \multicolumn{1}{c}{Ours} \\
\midrule
IS                         & 1.35                       & 1.28                       & 1.34                        & 1.37                         & 1.44                      & 1.52                        & \textbf{1.67}            \\
IL                         & 0.19                       & 0.21                       & 0.26                        & 0.17                         & 0.23                      & 0.29                        & \textbf{0.33}    \\
\bottomrule
\end{tabular}
}

\caption{Quantitative comparison of IS and IL on Visa.}
\label{tab:all_ISIL_visa}
\end{table}

\noindent\textbf{Qualitative Comparison on More Complex Categories.}~It is evident that our method can generate diverse and realistic anomaly data even for more complex categories, including multi-instance capsule, candle, and the more intricate PCB category. In comparison, it outperforms the best few-shot method, AnoDiff, as well as the best zero-shot method, RealNet.\\
\noindent\textbf{Downstream Task Evaluation: Image-Level Detection and Pixel-Level Localization.}~To validate the generalization capability of anomaly synthesis methods, we conduct comprehensive evaluations on both the MVTec AD and Visa datasets. For the Visa dataset, we synthesize 500 anomaly image-mask pairs per object category and train a U-Net model for downstream task evaluation. For the MVTec AD dataset, which provides fine-grained category annotations, we further synthesize 500 image-mask pairs for each anomaly type within every object category to verify robustness in complex scenarios.~Experimental results (as shown in Table~\ref{tab:all_visa} and Figure~\ref{tab:all_mvtec}) demonstrate that:~(1)On the Visa dataset, our method significantly outperforms existing approaches without requiring explicit category annotations, exhibiting strong adaptability to unseen classes.~(2)On the MVTec AD dataset, our approach surpasses all zero-shot baselines and even exceeds several 2023-2024 few-shot SOTA methods (e.g., DFMGAN~\cite{duan2023few} and AnoGen~\cite{gui2024anogen}) in specific metrics.~This comparative analysis confirms the stability and generalizability of our method across datasets and tasks, particularly aligning with practical industrial inspection requirements where prior knowledge is typically limited.

\begin{table}[]
\renewcommand{\arraystretch}{0.95} %
    \setlength{\tabcolsep}{2pt} %
    \resizebox{\linewidth}{!}{%
\begin{tabular}{l|cccccc}
\toprule
Method   & AUC-I         & AP-I          & F1-I          & AUC-P         & AP-P          & F1-P          \\
\midrule
DFMGAN   & 82.4          & 81.7          & 78.8          & 90.8          & 29.5          & 34.0          \\
AnoGEN   & 88.3          & 86.8          & 81.6          & 91.2          & 32.9          & 37.8          \\
AnoDiff  & 89.2          & 87.9          & 82.8          & 92.5          & 34.3          & 38.9          \\
CutPaste & 84.7          & 83.5          & 79.5          & 89.5          & 28.0          & 33.7          \\
DRAEM    & 86.1          & 84.7          & 81.2          & 90.4          & 29.7          & 34.9          \\
RealNet  & 90.2          & 89.3          & 84.4          & 92.8          & 34.9          & 39.4          \\
Ours     & \textbf{93.9} & \textbf{92.8} & \textbf{88.3} & \textbf{96.2} & \textbf{40.7} & \textbf{45.3} \\
\bottomrule
\end{tabular}
}
\caption{Performance comparison of anomaly data synthesized by different methods on VISA downstream detection tasks.}
\label{tab:all_visa}
\end{table}

\begin{table}[]
\renewcommand{\arraystretch}{0.95} %
    \setlength{\tabcolsep}{2pt} %
    \resizebox{\linewidth}{!}{%
\begin{tabular}{l|cccccc}
\toprule
Method   & AUC-I         & AP-I          & F1-I          & AUC-P         & AP-P          & F1-P          \\
\midrule
DFMGAN   & 87.2          & 94.8          & 94.7          & 90.0          & 62.7          & 62.1          \\
AnoGEN   & 97.7          & 98.8          & 97.4          & 97.9          & 75.3          & 69.0          \\
AnoDiff  & \textbf{99.2} & \textbf{99.7} & \textbf{98.7} & \textbf{99.1} & \textbf{81.4} & \textbf{76.3} \\
CutPaste & 75.8          & 89.9          & 88.4          & 91.7          & 52.9          & 51.4          \\
DRAEM    & 94.6          & 97.0          & 94.4          & 92.2          & 62.7          & 53.1          \\
RealNet  & 95.3          & 97.3          & 94.7          & 95.5          & 67.4          & 64.0          \\
Ours     & 97.6          & 98.3          & 97.3          & 98.3          & 74.9          & 67.9 \\
\bottomrule
\end{tabular}
}
\caption{Performance comparison of anomaly data synthesized by different methods on MVTec AD  downstream detection tasks.}
\label{tab:all_mvtec}
\end{table}

\section{More t-SNE result in Visa dataset}
\begin{figure*}[p]
  \centering
  \includegraphics[width=0.9\linewidth]{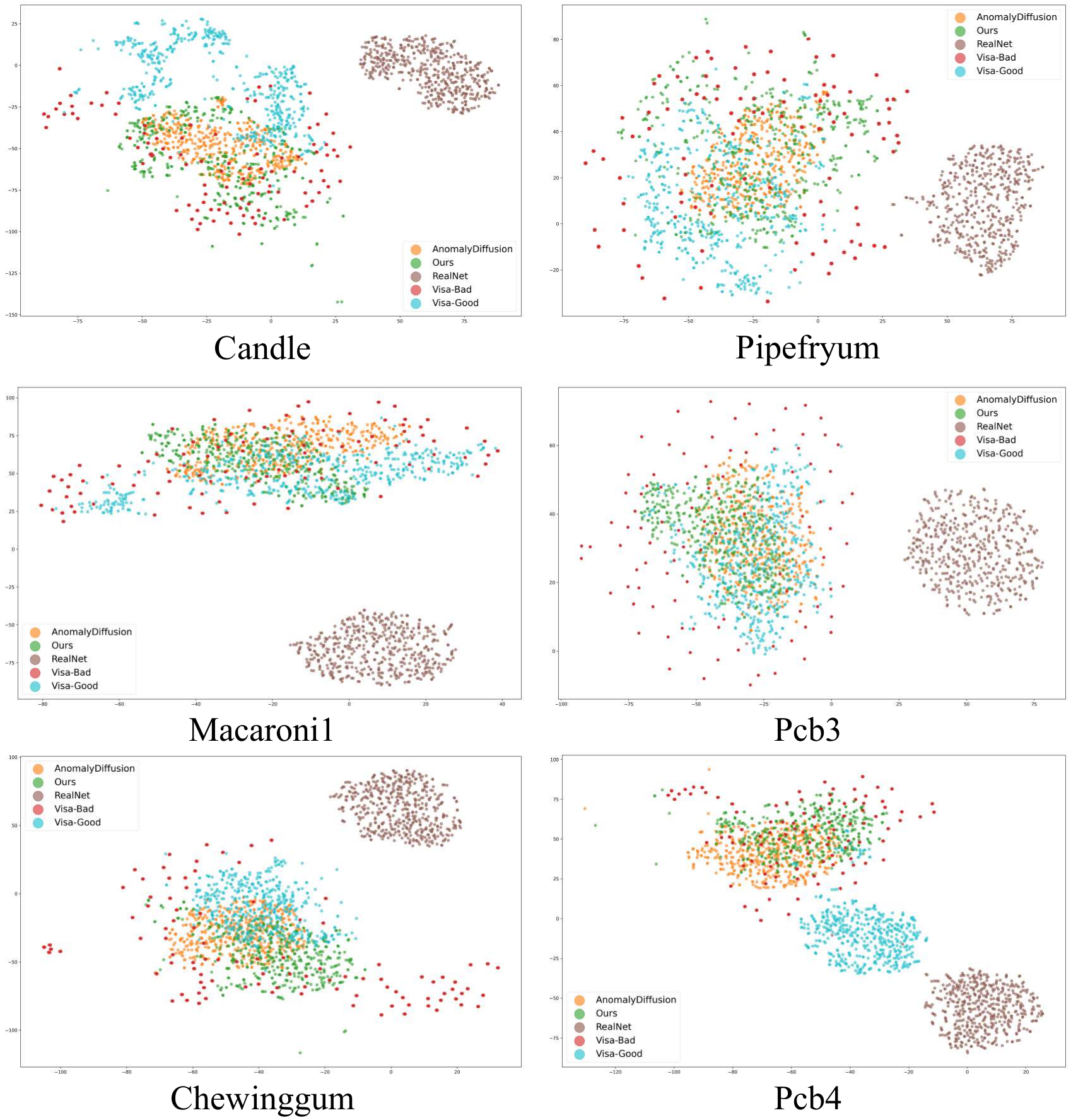}
  \caption{
    More t-SNE visualization examples of industrial objects in Visa.
  }
  \label{fig.13}
\end{figure*}

To comprehensively analyze the distribution of synthesized anomaly data from different methods, we apply t-SNE~\cite{van2008visualizing} to 500 anomaly images generated by RealNet, AnomalyDiffusion, and our method on the Visa dataset. Additionally, we select 500 normal industrial images and 500 real anomaly images (by replicating the 100 available real images for each object to reach 500). In total, we have five categories of data, with 500 images per category, which are reduced to 5x500 points for visualization.
In the main text, the 500 points for each method are further clustered into 100 points using k-means for visual clarity. 
Here we showcase six other examples in Visa, each fully displaying the situation of 500 points of five categories in Figure~\ref{fig.13}.

\begin{figure*}[p]
  \centering
  \includegraphics[width=0.85\linewidth]{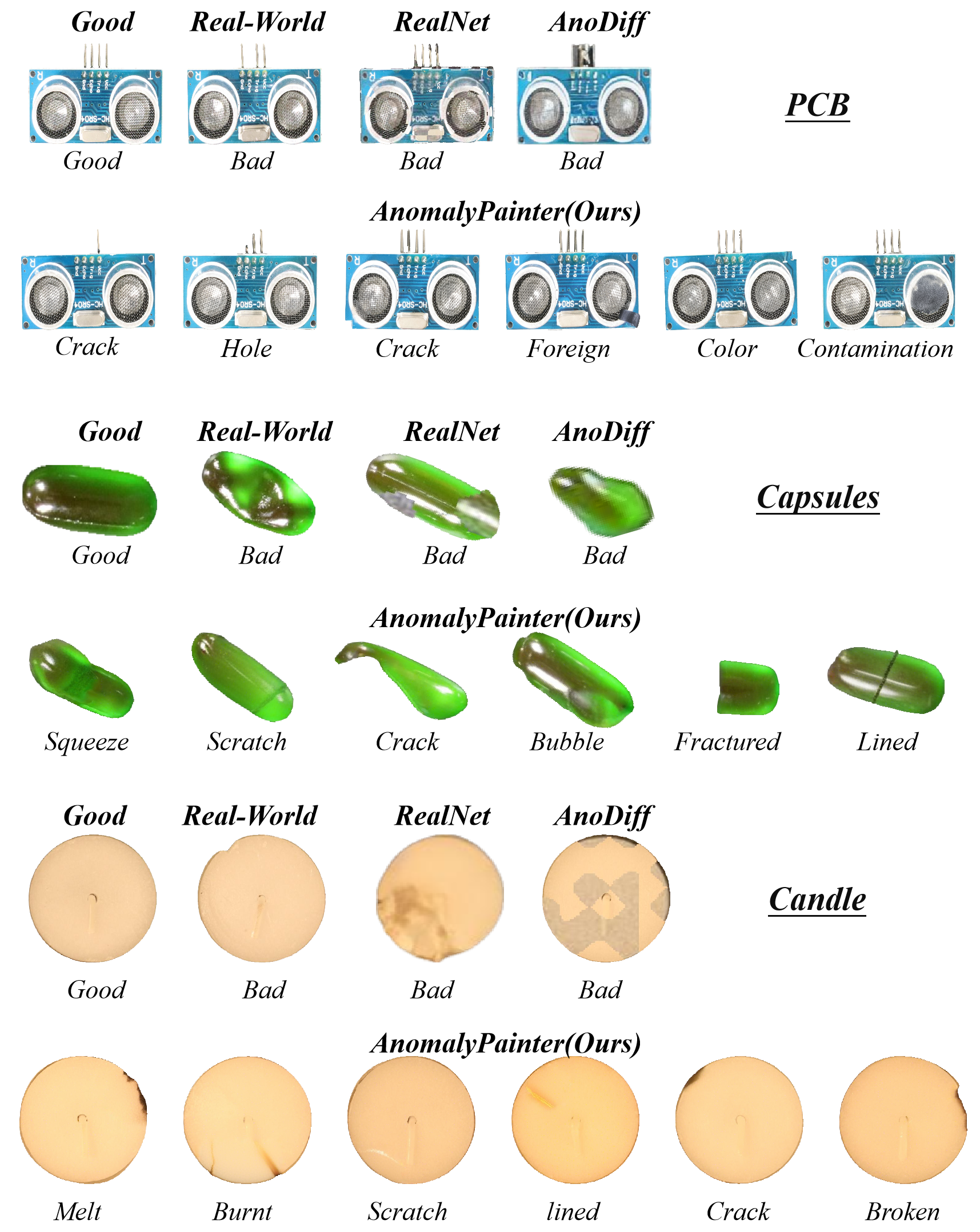}
  \caption{
    Qualitative comparison on complex categories.~Our method generates diverse and realistic anomalies for multi-instance capsule, candle, and complex PCB, outperforming AnoDiff and RealNet.
  }
  \label{fig.14}
\end{figure*}

\end{document}